\documentclass[letterpaper,10pt,conference]{ieeeconf}

\IEEEoverridecommandlockouts
\AtBeginDocument{
    \setlength{\abovedisplayskip}{4pt plus 1pt minus 1pt}
    \setlength{\belowdisplayskip}{4pt plus 1pt minus 1pt}
    \setlength{\abovedisplayshortskip}{3pt plus 1pt minus 1pt}
    \setlength{\belowdisplayshortskip}{3pt plus 1pt minus 1pt}
}

\usepackage{amsmath,amssymb}
\usepackage{graphicx}
\usepackage{booktabs}
\usepackage{multirow}
\usepackage{url}
\usepackage{titlesec}
\usepackage{xcolor}
\usepackage[colorlinks=true, linkcolor=blue, citecolor=blue, urlcolor=blue]{hyperref}

\title{\LARGE\bf
Dressing in Motion: A Human Motion-Aware Diffusion Policy for Robot-Assisted Dressing
}

\author{
Haoxiang~Sun$^{1}$,
Fangyuan~Wang$^{1}$,
Songhao~Huang$^{1}$,
Justina~Y.~W.~Liu$^{2}$,\\
Jihong~Zhu$^{3}$,
Peng~Zhou$^{4}$,
and David~Navarro-Alarcon$^{1}$%
\thanks{$^{1}$Department of Mechanical Engineering,
The Hong Kong Polytechnic University, Hong Kong SAR, China.
$^{2}$School of Nursing,
The Hong Kong Polytechnic University, Hong Kong SAR, China.
$^{3}$ School of Physics, Engineering and Technology, University of York, UK.
$^{4}$School of Advanced Engineering, 
Great Bay University, Dongguan, Guangdong, China.
Corresponding author: \texttt{dna@ieee.org}}
}
\begin{document}

\maketitle

\thispagestyle{empty}
\pagestyle{empty}

\begin{abstract}
Robotic dressing assistance is a promising solution for supporting older adults with physical impairments in daily living. However, dressing under human motion remains challenging, as complex garment--human contact and occlusions make it difficult to generate actions aligned with arm movements. In this letter, we propose a visuomotor policy that learns dressing skills from static expert demonstrations and generalizes to dynamic user-motion scenarios. A diffusion policy tailored to garment--human interaction geometry learns from partially observed point clouds with varied arm postures. We then introduce an object-centric representation based on PDE diffusion to capture the axial distribution of the arm. By sampling motion-relevant regions and registering them across consecutive observations, the proposed method approximates arm motion and reactively adapts the executed trajectory. We evaluate our method in simulation and a real-world human study involving nine participants, three garment types, and six arm-motion patterns. Results show that our method outperforms baselines in dressing progress, freedom of movement, and user comfort. The project website is \url{https://anonymous.4open.science/w/dressing-in-motion/}.
\end{abstract}


\section{Introduction}
Dressing is an essential component of activities of daily living (ADLs). Among ADLs, dressing is a physically intensive and repetitive task involving frequent body interaction. For older adults and individuals with physical impairments, dressing often requires assistance from professional caregivers. However, global population aging has intensified the shortage of trained caregivers, making robotic dressing a promising solution for autonomous assistance. In practical scenarios, users may naturally adjust their arm posture due to fatigue or discomfort, requiring the robot to adapt to human motion during assistance. In this work, we study a human motion-aware policy for robotic dressing to support a less restrictive and more user-accommodating assistance process.

In the context of garment manipulation, robotic dressing assistance remains particularly challenging. On the one hand, fabrics are deformable objects that lack compact state representations, and deformation estimation is difficult due to the infinite degrees of freedom and nonlinear dynamics. On the other hand, compared with conventional fabric manipulation tasks such as garment grasping \cite{zhuGuidingRoboticCloth2026} and folding \cite{zhangVisualTactileLearningGarment2023}, dressing is more contact-rich and interaction-intensive, as the garment usually needs to slide closely along the body surface. Unpredictable occlusions substantially reduce the accuracy of sleeve positioning. Meanwhile, excessive garment entanglement, tension, or potential contact between the gripper and human body may cause discomfort and even safety risks.

\begin{figure}[t]
    \centering
    \includegraphics[width=0.88\columnwidth, keepaspectratio]{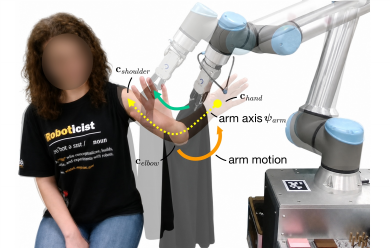}
    \caption{Robot-assisted dressing under human arm motion. The robot advances the garment along the arm axis while adapting to time-varying arm geometry.}
    \label{fig:1}
\end{figure}

Many prior studies on robot-assisted dressing have made substantial progress by using vision and/or force feedback to explicitly estimate human--garment contact states or implicitly learn action distributions \cite{wangOnePolicyDress2023,kotsovolisModelPredictiveControl2024}. In parallel with these advances, reliable dressing behaviors remain difficult to fully specify through low-dimensional state and under-parameterized reward functions, making learning from physical demonstrations a more natural paradigm for acquiring dressing skills. Recent advances in data-driven learning have shown great potential for deformable object manipulation, with representative examples including robotic foundation models \cite{intelligencePi07SteerableGeneralist2026} and generative AI models \cite{chiDiffusionPolicyVisuomotor2024} that learn transferable policy distributions from high-quality expert datasets.

However, most existing robotic dressing approaches are conducted under a static-arm assumption, where the subject is required to keep the arm fixed throughout the dressing process \cite{pignatLearningAdaptiveDressing2017, ericksonDeepHapticModel2018, Sun_ral2024, kotsovolisGarmentDiffusionModels2025}. This assumption provides a strong prior for handling visual occlusions, since the human-skeleton topology, including the hand, elbow, and shoulder \cite{zhuYouNeedHand2024}, or even the full arm point-cloud distribution \cite{zhangProbabilisticRealTimeUser2019}, can be inferred from the initial configuration in some settings. Nevertheless, this simplification inevitably compromises the user experience. In an ideal dressing process, users should be allowed to freely adjust their arm posture when they feel fatigue or need to move, while the robot should compliantly adapt to human motion and complete the dressing task accordingly.

In this work, we address the problem of robotic dressing under human arm motion. Our key idea is to combine a visuomotor diffusion policy learned from static-arm demonstrations with a real-time adaptive control strategy for dynamic human postures. The proposed dressing diffusion policy is designed to be sensitive to human--garment interaction geometry by utilizing partially observed point-cloud inputs. We introduce an object-centric local-frame representation \cite{bilalogluObjectcentricTaskRepresentation2026} for the arm surface and use it to automatically sample motion-relevant regions from the arm point cloud. These sampled regions are registered online to estimate user motion and project the learned dressing actions, enabling the policy to adapt to human dynamics during execution. We note that comparing visual and force-based sensing modalities is not the focus of this paper, while incorporating force feedback can serve as a future extension of the proposed method.

The contributions of this paper are as follows:  
\begin{itemize}
    \item A diffusion policy tailored to garment--human interaction geometry for robotic dressing.
    \item A PDE-diffused scalar field that forms an arm-axis representation to localize motion-relevant regions from partially observed arm point clouds.
    \item An online trajectory adaptation scheme that estimates arm motion through point registration and projects actions for motion-compatible execution.
    \item Comprehensive simulation and real-world evaluations across diverse garment types and human motion patterns, demonstrating improved dressing progress, motion tolerance, and user ergonomics.
\end{itemize}

\section{Related Work}
\subsection{Robot-Assisted Dressing}
Robot-assisted dressing has received considerable research attention. Early works studied user modeling to enable personalized dressing \cite{zhangProbabilisticRealTimeUser2019,gaoUserModellingPersonalised2015, chanceElbowsOutPredictiveTracking2018}. Subsequent studies explored the incorporation of visual observations and/or force feedback as inputs, either by using model-based methods to learn garment--human dynamics and integrate them with control strategies \cite{kotsovolisGarmentDiffusionModels2025, tsukakoshiCloseFittingDressingAssistance2026}, or by directly inferring action policies within reinforcement learning \cite{wangOnePolicyDress2023, Sun_ral2024, cleggLearningCollaborateSimulation2020} and imitation learning frameworks \cite{blancoEvaluatingPreDressingStep2025, franzeseGeneralizableMotionPolicies2025, zhaoBimanualRobotAssistedDressing2025}. However, these approaches invariably rely on the common assumption that the human arm remains static during dressing. While this assumption simplifies the inference of human-skeleton topology under occlusion, it also limits user comfort and natural interaction. More recent work \cite{Hao_corl2025} has begun to address dressing assistance for humans with arm motion in unstructured scenarios, by fine-tuning a simulation-trained reinforcement learning policy with real-world force feedback. However, real-time adaptation to human motion remains limited. Our work differs by contributing an imitation learning framework that learns from static demonstrations while maintaining online awareness of arm-motion variability during execution.

\subsection{Learning Deformable Object Manipulation Skills}
Dressing is inherently a contact-rich deformable object manipulation (DOM) task involving close human--robot interaction, where accurate state estimation is challenging due to the nonlinear dynamics and infinite degrees of freedom of deformable materials. For this human-mastered skill, reinforcement learning methods require faithful digital twins and reliable sim-to-real transfer, both remain challenging in practice. In contrast, learning from expert demonstrations provides a promising paradigm for acquiring action policies while bypassing explicit physical modeling. Learning manipulation skills from demonstrations has evolved from model-based regression \cite{zhuLearningTaskParameterizedSkills2022} and behavior cloning \cite{florenceSelfSupervisedCorrespondenceVisuomotor2020} to generative modeling approaches \cite{wengDexDiffuserGeneratingDexterous2024, zhangAffordancebasedRobotManipulation2025}, which have demonstrated strong capability in learning complex data distributions. Among them, diffusion models have emerged as a powerful tool for constructing visuomotor policies that capture multimodal action distributions in DOM \cite{chiDiffusionPolicyVisuomotor2024}, including tissue shape control \cite{scheiklMovementPrimitiveDiffusion2024} and fabric folding \cite{molettaPreferenceAlignedVisuomotor2026}.

In this work, we focus on robot-assisted dressing with human motion, an object-centric DOM task that involves physical interaction with the arm surface and requires adaptation to variations in the spatial position and local curvature of the contact surface \cite{longhiniUnfoldingLiteratureReview2025}. Improper contact can lead to entanglement around the hand and elbow. However, prior work studied shape control of deformable objects in structured environments, where environmental contact is typically static or ignored. In addition, such physical interaction results in cluttered visual observations under occlusion. For this reason, our approach is tailored to garment--human interaction geometry while modeling the target distribution of desired dressing actions under the guidance of arm-motion topology.

\section{Problem Formulation and Assumptions}
We formulate robotic dressing under human arm motion as a visuomotor policy learning problem modeled as a partially observable Markov decision process (POMDP). Let $\mathbf{c}_{\text{sleeve},t}$ and $\mathbf{c}_{\text{hand},t}$ denote the sleeve-opening center and hand position at time $t$, respectively, and let $\boldsymbol{\psi}_{\text{arm},t}$ denote the curvilinear arm centerline extending from the hand toward the shoulder through the medial axis of the approximately cylindrical arm geometry, as shown in Fig.~\ref{fig:1}. Differing from approaches that begin with the arm already inserted into the sleeve, the robot starts from an initial pose while grasping the garment in front of the arm, without requiring pre-alignment between $\mathbf{c}_{\text{sleeve},t}$ and $\mathbf{c}_{\text{hand},t}$. It first reduces the hand--sleeve alignment error $\|\mathbf{c}_{\text{sleeve},t}-\mathbf{c}_{\text{hand},t}\|$
and orients the sleeve for insertion onto the forearm, thereby resolving the alignment and orientation constraints during the pre-insertion phase.

The task objective is to progressively advance the garment sleeve over the hand and elbow until it reaches the shoulder while accommodating time-varying arm geometry. Given a partial point cloud $\mathbf{P}_t$, the goal is to learn a policy $\pi_{\theta}$ that maps the observation embedding $\mathbf{o}_t$, constructed from $\mathbf{P}_t$ and the robot state, to an action distribution
$\mathbf{a}_t\sim\pi_{\theta}(\cdot\mid\mathbf{o}_t)$.
Let
$\mathcal{Y}_{h,t}
=
\{\mathbf{c}_{\text{shoulder},t},
\mathbf{c}_{\text{elbow},t},
\mathbf{c}_{\text{hand},t}\}$
denote the arm landmarks, and let $\boldsymbol{\chi}_{h,t}$ denote the arm joint configuration. The arm pose is represented as
$\mathbf{q}_{h,t}=(\mathcal{Y}_{h,t},\boldsymbol{\chi}_{h,t})$, and
$t_{\mathrm{ins}}$ denotes the sleeve insertion time. During the pre-insertion phase, the arm remains static, such that
$\mathbf{q}_{h,t}=\mathbf{q}_{h,0}$ for $t<t_{\mathrm{ins}}$.
After insertion, the participant may vary $\mathbf{q}_{h,t}$ for
$t\geq t_{\mathrm{ins}}$, resulting in time-varying arm geometry during garment advancement.

\begin{figure*}[!t]
    \centering
    \includegraphics[width=0.95\textwidth, keepaspectratio]{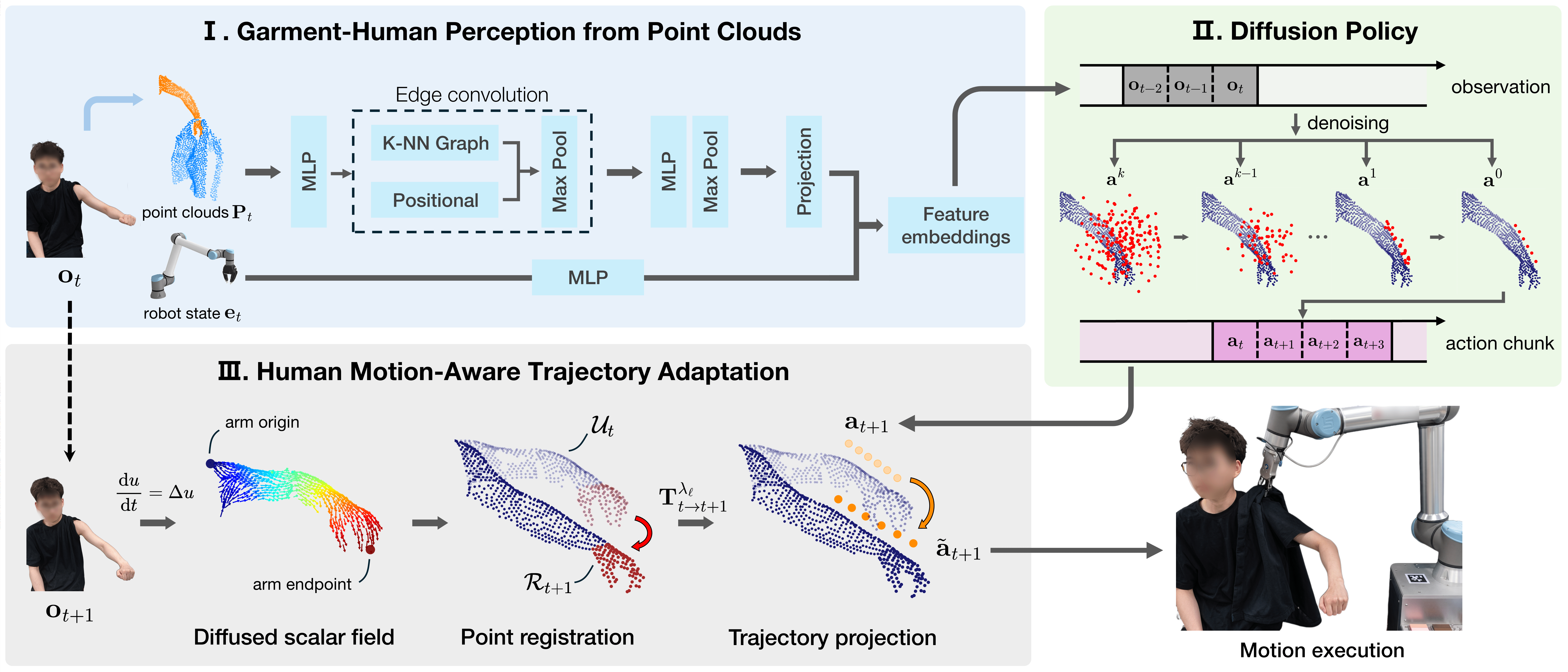}
    \caption{Overview of the proposed method. \textit{Top left}: Visual encoder extracts point-cloud features and fuses them with the robot state to obtain the observation $\mathbf{o}_t$. \textit{Top right}: A diffusion policy denoise Gaussian noise to predict an action chunk. \textit{Bottom}: Diffused scalar field identifies the region $\mathcal{R}_{t+1}$, which is registered to the reference region $\mathcal{R}_{t}$ to estimate arm motion. The estimated transformation is applied to generate the adapted execution trajectory.}
    \label{fig:2}
\end{figure*}

\section{Method}
\textit{Method Overview:} 
As shown in Fig.~\ref{fig:2}, our method follows a hierarchical framework. At the high level, a diffusion model learns a general motion policy from static demonstrations covering variations in arm poses and garment types. To further address dynamic dressing, we construct a diffused scalar field from occluded point clouds to obtain a topologically continuous coordinate along the arm axis, and perform point registration to estimate transformations of the arm pose, which are then used to align the generated motion with the current arm configuration and reactively respond to non-Markovian arm motions. During execution, the diffusion policy updates action chunks at a lower frequency, while trajectory adaptation (TA) runs at a higher rate to generate end-effector commands.

\subsection{Diffusion Policy Learning from Demonstrations}
The first core of this work is to learn a visuomotor policy for robot-assisted dressing from expert demonstrations. Visual observations in this task inevitably contain entanglement and occlusions between the garment and the arm. We introduce the dressing diffusion policy, which learns a distribution over multimodal actions by training a diffusion policy that iteratively denoises actions $\mathbf{a}$ conditioned on occluded visual features $\mathbf{s}$. For the observation space $\mathcal{O}$, let $\mathcal{P}$ denote the human--garment point space, and $\mathbf{P}_t\subset\mathcal{P}$ is the arm--garment point set cropped at time $t$ from a rectangular region surrounding the dressing-side arm. The policy observation $\mathbf{o}_t$ is defined as a latent feature that fuses the point-cloud feature extracted from $\mathbf{P}_t$ with the encoded end-effector pose $\mathbf{e}_t$. The points $\mathbf{P}_t$ are extracted from a rectangular region around the dressing-side arm and contains a mixture of human and garment points. At each time step $t$, the points are downsampled by farthest point sampling (FPS) to preserve spatial coverage with a fixed point-set size, without additional segmentation.

To encode human--garment interaction geometry and extract fine-grained 3D features, the sampled points are processed by a lightweight encoder consisting of an EdgeConv layer \cite{dgcnn}. For each point $\mathbf{p}_i$, the EdgeConv feature is computed as:
\begin{equation}
\mathbf{h}_i = \max_{j\in\mathcal{N}_K(i)}
\phi_{\eta}
\left(
\left[
\mathbf{p}_i,
\mathbf{p}_j-\mathbf{p}_i
\right]
\right)
\end{equation}
where $\mathcal{N}_K(i)$ denotes the $K$-nearest-neighbor set of $\mathbf{p}_i$ in the sampled point cloud, and $\phi_{\eta}$ is a shared MLP applied to each edge feature. This operation combines the point position $\mathbf{p}_i$ with local offsets $\mathbf{p}_j-\mathbf{p}_i$ over the neighborhood graph to capture local interaction geometry between the human arm and garment. After max aggregation over neighboring edge features, the EdgeConv layer produces a point-feature tensor of shape $N\times d_h$, where $N$ is the number of sampled points and $d_h$ is the output feature dimension. The resulting point-cloud feature is projected and fused with the encoded end-effector pose feature to obtain $\mathbf{o}_t$, which conditions a diffusion model that iteratively denoises Gaussian noise $\mathbf{a}^{k}$ into actions $\mathbf{a}^0$:
\begin{equation}
    \mathbf{a}^{k-1} = \sqrt{\bar\alpha^{k-1}}\, \hat{\mathbf{a}}_\theta(\mathbf{o}_t,\mathbf{a}^k,k)
+ \sqrt{1\!-\!\bar\alpha^{k-1}}\, \mathbf\epsilon_\theta(\mathbf{o}_t,\mathbf{a}^k,k)
\end{equation}
where $\mathbf\epsilon_\theta$ is the noise prediction network and 
\begin{equation}
\hat{\mathbf{a}}_\theta = \frac{
\mathbf{a}^k - \sqrt{1-\bar\alpha^k}\,\mathbf\epsilon_\theta(\mathbf{o}_t,\mathbf{a}^k,k)
}{
\sqrt{\bar\alpha^k}
}.
\end{equation}
Here, DDIM \cite{song2020denoising} is used as the sampling scheduler for the reverse denoising process. The term $\bar\alpha^k=\prod_{s=1}^{k}\alpha_s$ denotes the cumulative product of the predefined variance schedule. The network is trained with the standard noise-prediction objective:
\begin{equation}
    \mathcal{L}(\theta)
    =
    \mathbb{E}_{\mathbf{a}_0,\mathbf{o}_t,k,\mathbf{\epsilon}}
    \left[
    \left\|
    \mathbf{\epsilon}
    -
    \mathbf{\epsilon}_\theta(\mathbf{o}_t,\mathbf{a}^k,k)
    \right\|_2^2
    \right].
\end{equation}

\subsection{Arm-Centric Diffused Scalar Field}
The diffusion model outputs each action chunk over a prediction horizon ${T}_p=8$ at 15 Hz for downstream processing. During dressing, the subject is allowed to adjust the arm pose, which requires the policy to respond at over 50 Hz with an action execution horizon ${T}_a=4$. Unlike prior approaches that reconstruct the full arm geometry under occlusion, which is not assumed under changing arm configurations. Instead, we extract an effective region from partially observed point clouds for motion adaptation. This is based on the practical premise that the dressing outcome is mainly governed by the interaction between the robot and the uncovered arm region $\mathcal{U}\subset\mathcal{P}$. In particular, the region of interest $\mathcal{R}\subset\mathcal{U}$ is defined as the outermost uncovered arm points closest to the current garment boundary along the arm axis $\boldsymbol{\psi}_{\text{arm},t}$, which largely determines the sleeve--arm insertion behavior. Guided by this premise, the motion-adaptation problem is reformulated as identifying $\mathcal{R}$ from cluttered point clouds and using it to approximate inter-frame arm motion.

Specifically, accurately sampling $\mathcal{R}$ from occluded visual observations entangled with multiple objects is nontrivial. For this reason, we construct a scalar field to obtain a geometry-aware axial distribution of the observed arm surface by introducing a point sampling method based on PDE diffusion \cite{bilalogluObjectcentricTaskRepresentation2026}. The scalar field $u(\cdot,\tau):\mathcal{P}\rightarrow\mathbb{R}$ is defined over the observed points, where $\tau$ denotes the diffusion time. The smoothness of $u$ follows the Dirichlet energy principle, which penalizes large spatial gradients and encourages neighboring arm-surface points to have consistent scalar values. In the continuous setting, the energy minimizer under boundary conditions satisfies Laplace's equation:
\begin{equation}
\Delta u(\mathbf{x}) = 0, \quad \mathbf{x}\in\Omega
\end{equation}
where $\Omega$ denotes the underlying arm surface. The gradient flow of this energy yields the continuous diffusion process, which is discretized on the observed point cloud using the point-cloud Laplacian:
\begin{equation}
\partial_{\tau} u(\mathbf{p},\tau) = \Delta_{\mathcal{P}} u(\mathbf{p},\tau),\  \ \mathbf{p} \in \mathcal{P}
\end{equation}
where $\Delta_{\mathcal{P}}$ denotes the point-cloud Laplacian. To propagate the local axial distribution of the arm, we specify the shoulder vertex as a source keypoint $\mathbf{s}$ acting as a physical repulsor, with the initial condition $u(\mathbf{s},0)=1$ and $u(\mathbf{p},0)=0$ for all $\mathbf{p}\in\mathcal{P}\setminus{\mathbf{s}}$. The diffusion equation is solved by implicit time stepping as:
\begin{equation}
    \mathbf{u}_\tau = (\mathbf{M}-\tau\mathbf{C})^{-1}\mathbf{M} \mathbf{u}_0
\end{equation}
where $\mathbf{u}_{0},\mathbf{u}_{\tau}\in\mathbb{R}^{|\mathcal{P}|}$ denote the point-wise scalar values before and after diffusion. $\mathbf{C}$ is the weak Laplacian matrix constructed from cotangent weights of local tangent-plane triangulations \cite{sharpLaplacianNonmanifoldTriangle2020}, and $\mathbf{M}$ is the diagonal mass matrix determined by the areas of triangles adjacent to each point.

\begin{figure}[t]
    \centering
    \includegraphics[width=0.95\columnwidth, keepaspectratio]{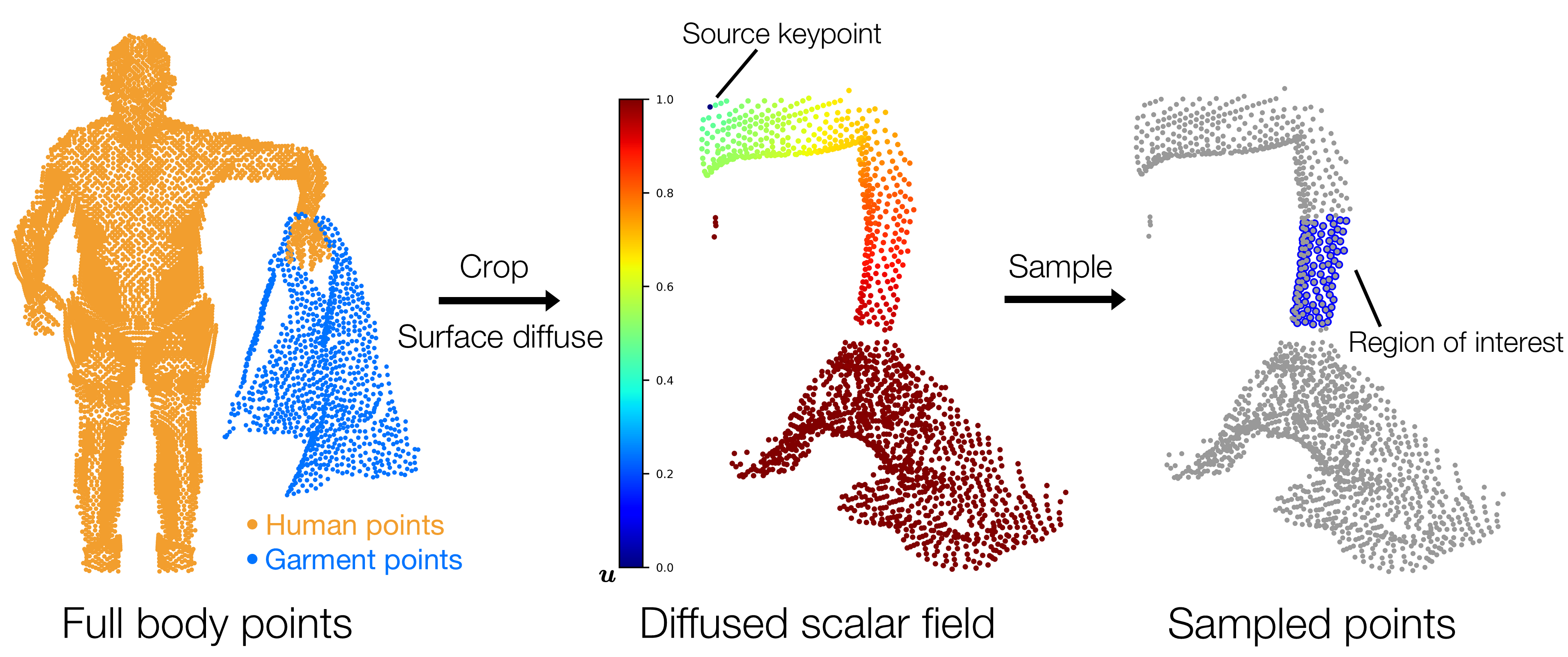}
        \caption{Diffused scalar-field construction and ROI sampling. Surface diffusion assigns each arm point a scalar value $u$ encoding its geodesic position along the arm axis. The uncovered arm region $\mathcal{U}$ is extracted, from which the region of interest $\mathcal{R}$ is sampled.}
    \label{fig:3}
\end{figure}

The sampling procedure based on the diffused scalar field is illustrated in Fig.~\ref{fig:3}. We adjust $\tau$ to perform short-time diffusion, which yields a scalar field sensitive to geodesic distance. By imposing continuity and smoothness constraints on the diffused scalar field along the arm axis from the shoulder source vertex, we reject garment points and wrinkle-induced artifacts, yielding the uncovered arm region $\mathcal{U}$.

\subsection{Arm-Motion-Aware Trajectory Adaptation}

The action chunk
$\mathbf{A}_t=\{\mathbf{a}_{t},\dots,\mathbf{a}_{t+T_a-1}\}$
is predicted at time $t$ and contains desired end-effector
trajectories expressed in the arm-centric frame associated with $\mathbf{o}_t$.
At each execution step $\ell=0,\dots,T_a-1$, the diffused scalar field
is used to sample the current region of interest
$\mathcal{R}_{t+\ell}$. We then register the reference region $\mathcal{R}_t$, obtained at prediction time, to the current region $\mathcal{R}_{t+\ell}$ using generalized iterative closest point (GICP). Let $\mathcal{M}$ denote the set of correspondences between $\mathbf{p}_i^t\in\mathcal{R}_t$ and $\mathbf{p}_j^{t+\ell}\in\mathcal{R}_{t+\ell}$. For each $(i,j)\in\mathcal{M}$, the residual is defined as $\mathbf{r}_{ij}=\mathbf{p}_j^{t+\ell}-(\mathbf{R}\mathbf{p}_i^t+\boldsymbol{\delta})$. GICP estimates the inter-frame transformation by solving
\begin{equation}
\mathbf{T}_{t\rightarrow t+\ell}
\!=\!
\operatorname*{arg\,min}_{\mathbf{T}\in SE(3)}
\!\sum_{(i,j)\in\mathcal{M}}
\mathbf{r}_{ij}^{\top}
\left(
\boldsymbol{\Sigma}_{j}^{t+\ell}
\!+\!
\mathbf{R}\boldsymbol{\Sigma}_{i}^{t}\mathbf{R}^{\top}
\right)^{-1}
\!\mathbf{r}_{ij}
\end{equation}
where $\mathbf{T}=(\mathbf{R},\boldsymbol{\delta})$ comprises the rotation $\mathbf{R}\in SO(3)$ and translation $\boldsymbol{\delta}\in\mathbb{R}^{3}$, and $\boldsymbol{\Sigma}_{i}^{t}$ and $\boldsymbol{\Sigma}_{j}^{t+\ell}$ are the local covariance matrices of the matched points. Restricted to the motion-relevant arm region, $\mathbf{T}_{t\rightarrow t+\ell}$ provides a local rigid estimate of the inter-frame arm displacement. We use this transformation matrix to project each pending position target through a bounded correction scheme:
\begin{equation}
\begin{bmatrix}
\widetilde{\mathbf{d}}_{t+\ell}\\
1
\end{bmatrix}
=
\mathbf{T}^{\lambda_{\ell}}_{t\rightarrow t+\ell}
\begin{bmatrix}
\mathbf{d}_{t+\ell}\\
1
\end{bmatrix}
\end{equation}
where
\begin{equation}
\begin{aligned}
&\mathbf{T}^{\lambda_{\ell}}_{t\rightarrow t+\ell}
=
\operatorname{Exp}\!\left(
\lambda_{\ell}
\operatorname{Log}
\left(
\mathbf{T}_{t\rightarrow t+\ell}
\right)
\right)\\
&\lambda_{\ell}
=
\min\!\left(
1,
\frac{\rho}{
\left\|
\operatorname{Log}
\left(
\mathbf{T}_{t\rightarrow t+\ell}
\right)
\right\|
+\varepsilon}
\right).
\end{aligned}
\end{equation}
The projected action is
\begin{equation}
\widetilde{\mathbf{a}}_{t+\ell}
=
\begin{bmatrix}
\widetilde{\mathbf{d}}_{t+\ell}\\
\boldsymbol{\varphi}_{t+\ell}
\end{bmatrix},
\end{equation}
where $\mathbf{d}_{t+\ell}$ and $\boldsymbol{\varphi}_{t+\ell}$ denote the predicted nominal end-effector position and orientation, respectively. The scalar $\lambda_\ell$ bounds the correction magnitude by the hyperparameters $\rho$ and $\varepsilon$ to avoid abrupt motion updates. This projection adapts the translational component of the end-effector target according to the estimated local arm displacement while preserving the predicted orientation.

\section{Results}
\label{sec:results}

\subsection{Numerical Simulation}
\label{sec:numerical_simulation}
\begin{figure}[!t]
    \centering
    \includegraphics[width=0.92\columnwidth, keepaspectratio]{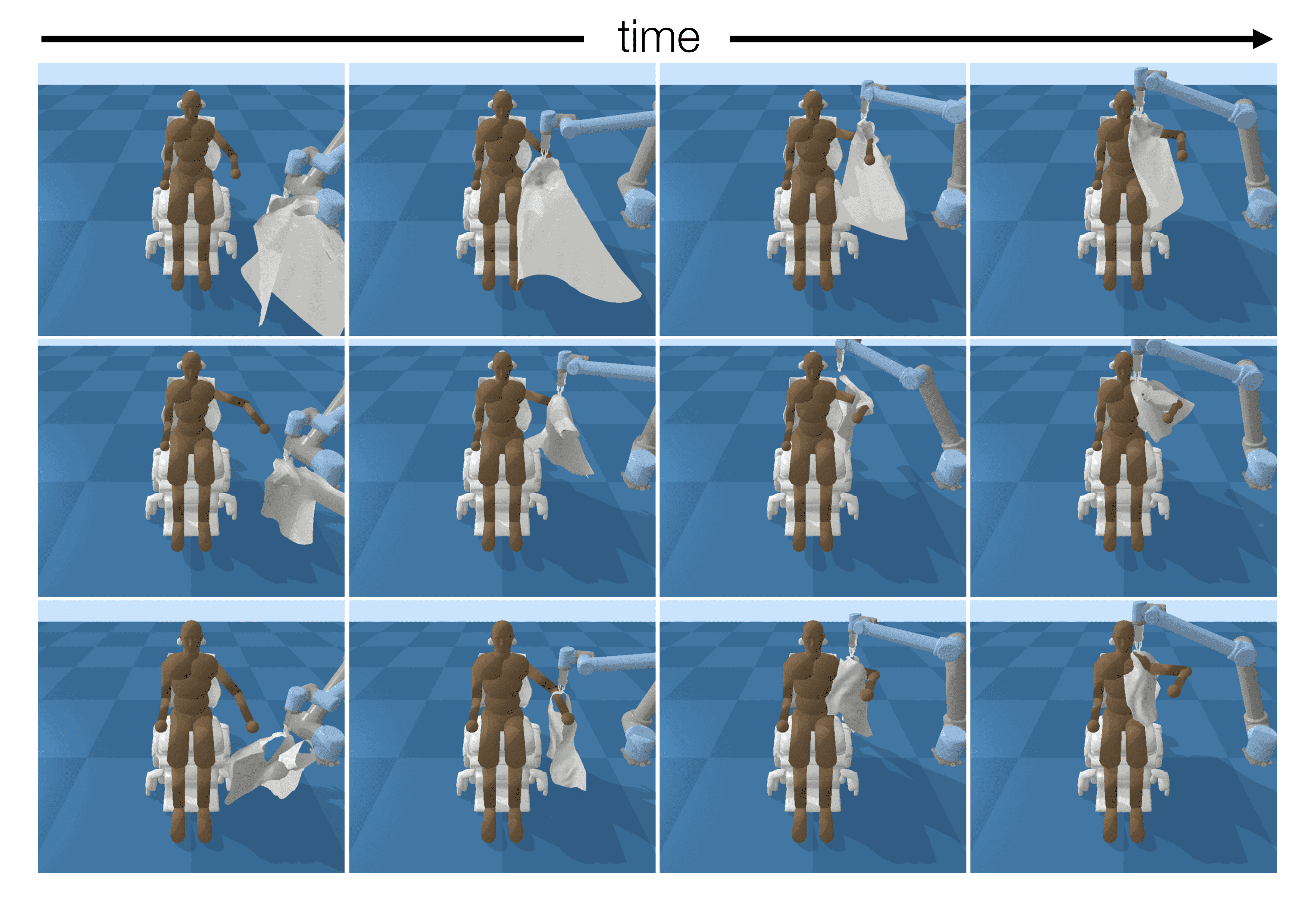}
    \caption{Examples of dressing process with different garment types and human models in the simulation environment.}
    \label{fig:4}
\end{figure}

We evaluate our method in the Assistive Gym simulation environment \cite{ericksonAssistiveGymPhysics2020}. Four human body models spanning genders and body-shape variations are configured, corresponding to male/female and slim/heavy morphologies. For each model, the arm posture is controlled by two revolute joints located at the shoulder and elbow, each with a motion range of $45^\circ$. Three garment types with different sleeve configurations are cropped from the Cloth3D dataset \cite{berticheCLOTH3DClothed3D2020} and imported into the simulator. Representative simulation trials are shown in Fig.~\ref{fig:4}.

In terms of arm motion, we evaluate both static and dynamic conditions in simulation by defining ten arm motion patterns. Eight motions are generated from planar arm displacements in different directions, while the remaining two correspond to clockwise and counterclockwise rotational arm movements. In addition, we evaluate three arm motion speed profiles determined by the joint angular velocity $v$: static motion with $v=0$, moderate-speed motion with $v=1.0~\mathrm{rad/s}$, and fast motion with $v=2.0~\mathrm{rad/s}$.

We compare the proposed method and its ablated variant against several baselines:
\begin{itemize}
  \item \textbf{Ours}: the full proposed framework.
  \item \textbf{Ours w/o TA}: an ablated variant of our method without the human motion-aware trajectory adaptation module.
  \item \textbf{DP3}~\cite{ze3DDiffusionPolicy2024}: a point-cloud-conditioned diffusion policy for general robotic manipulation, used to evaluate the effect of our interaction-aware point-cloud encoding.
  \item \textbf{Diff-MPC}~\cite{kotsovolisGarmentDiffusionModels2025}: a diffusion garment-dynamics MPC baseline with its original one-hot action encoding replaced by end-effector action feature encoding for fair comparison.
  \item \textbf{DP-image}~\cite{chiDiffusionPolicyVisuomotor2024}: an image-conditioned diffusion policy.
  \item \textbf{BC-LSTM}~\cite{mandlekarWhatMattersLearning2022}: a recurrent behavior cloning baseline that models temporal dependencies with an LSTM.
\end{itemize} 

\textbf{Training.} In training, all imitation learning methods are trained on the dataset containing 180 trajectories obtained through teleoperation in simulation. For Diff-MPC, we follow its original model-based reinforcement learning setting and train it with 750 trajectories. During evaluation, each trial is limited to a maximum of 500 execution steps. Performance is evaluated using two metrics: sleeve insertion success and dressing ratio, defined as the ratio of the dressed arm length to the total arm length. For each benchmark and arm-motion speed profile, we conduct 70 trials with randomized garment--human initializations. Results are reported across all combinations of arm motions, garment types, and human body models.

\begin{figure}[!t]
    \centering
    \includegraphics[width=0.95\columnwidth]{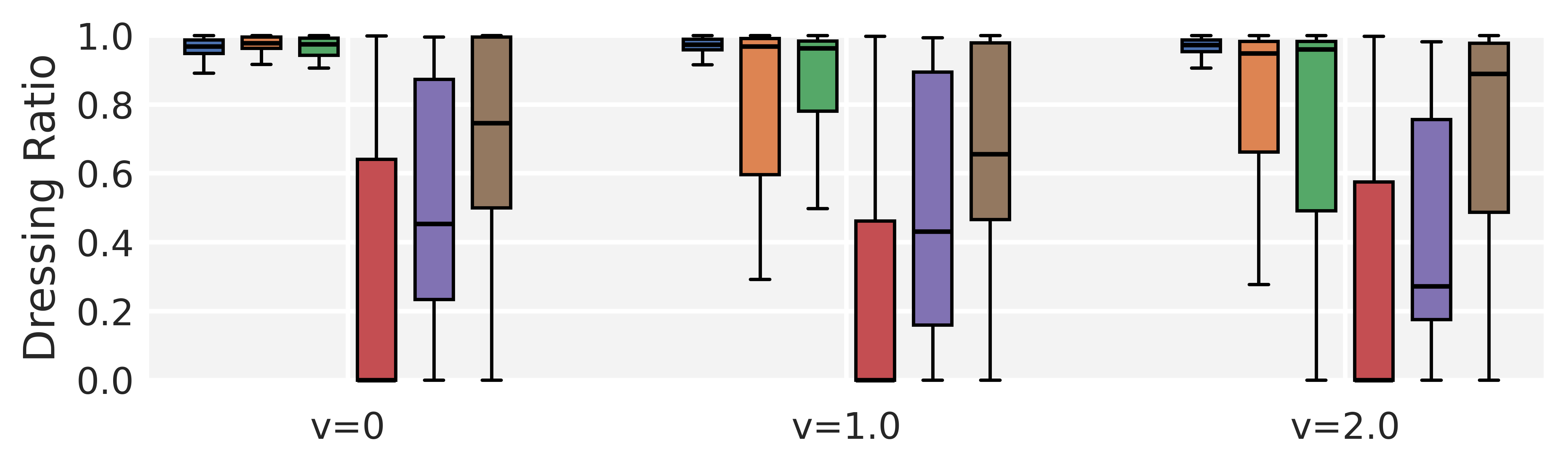}
    \includegraphics[width=0.95\columnwidth]{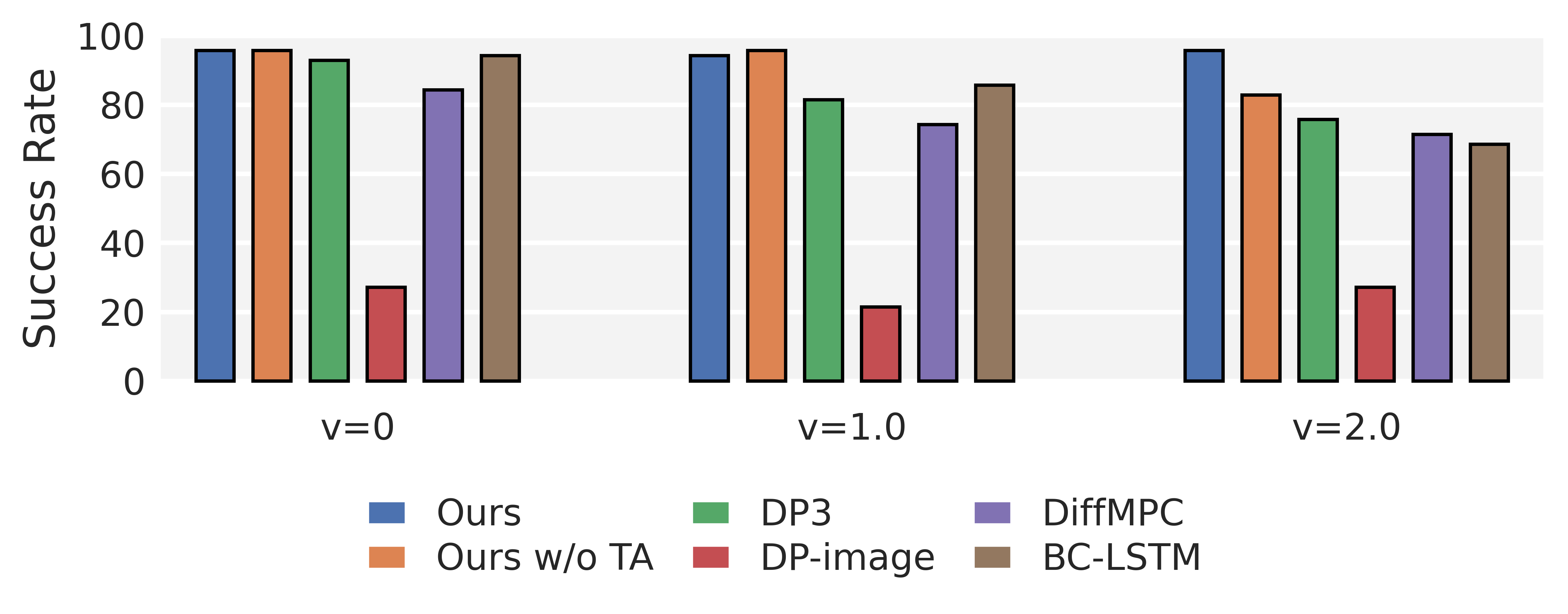}
    \caption{Simulation performance under different arm-motion speeds. For each benchmark and speed profile, results are computed over 70 randomized trials. \textit{Top}: dressing ratio. \textit{Bottom}: sleeve insertion success rate.}
    \label{fig:5}
\end{figure}

Fig.~\ref{fig:5} shows the simulation results over all 1470 trials. Under static conditions, several methods achieve reasonable performance, with point-cloud conditioned diffusion policies outperforming image-based policies by using richer 3D geometric cues. As arm motion is introduced and the motion speed increases, most baselines show degraded performance. Since these methods lack explicit motion adaptation, hand motion can pull the arm away from the garment opening during alignment, causing insertion failure or elbow entanglement. The comparison between Ours and Ours w/o TA further shows that trajectory adaptation provides consistent gains across different movement types, indicating its role in handling arm-motion variability. In addition, the comparison between Ours w/o TA and DP3 demonstrates the benefit of the proposed interaction-aware point-cloud encoding for generalization across garments and arm poses. The full proposed method remains robust to human motion, achieving an average dressing ratio and sleeve insertion success rate above 0.95.

\begin{figure}[!t]
    \centering
    \includegraphics[width=0.95\columnwidth, keepaspectratio]{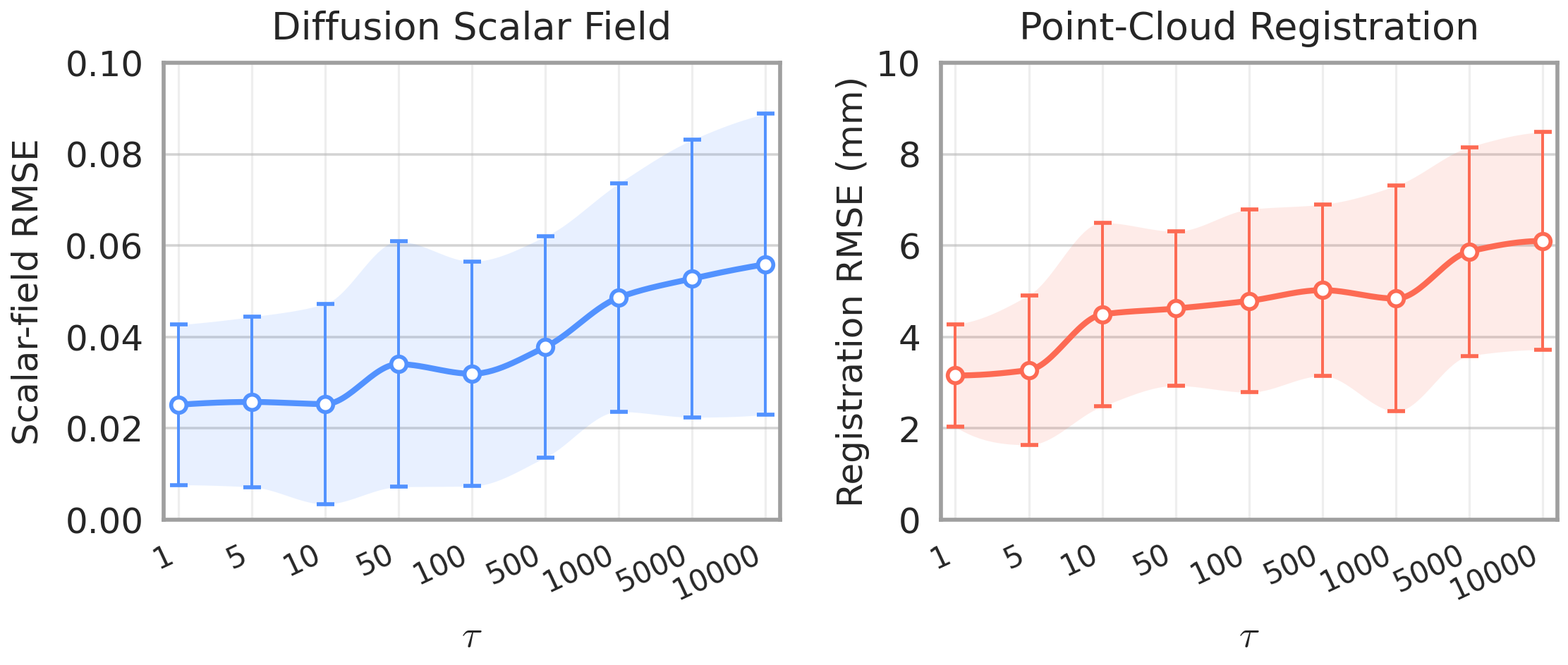}
    \caption{Error bar plots of the diffused scalar field and point-cloud registration under different diffusion times $\tau$. The plots report the mean RMSE over 50 independent randomized trials, with error bars indicating standard deviations.}
    \label{fig:6}
\end{figure}

\textbf{Diffusion-Time Robustness.} We evaluate the robustness of the diffused scalar field and point-cloud registration process under different diffusion times $\tau$, as shown in Fig.~\ref{fig:6}. The evaluation is conducted over 50 randomized trials. Scalar RMSE quantifies the alignment consistency of the rank-normalized diffusion scalar field by comparing matched current-reference arm points after registration. Registration RMSE measures geometric alignment error using the symmetric nearest-neighbor RMS distance between the registered current arm cloud and the reference arm cloud. In principle, increasing $\tau$ produces a smoother scalar field that is resilient to noise, whereas short-time diffusion better preserves gradient information along the arm axis. The mean RMSE results indicate that the proposed point sampling and registration process remains robust across diffusion times.

\subsection{Hardware Experiments}
\begin{figure}[!t]
    \centering
    \includegraphics[width=0.94\columnwidth, keepaspectratio]{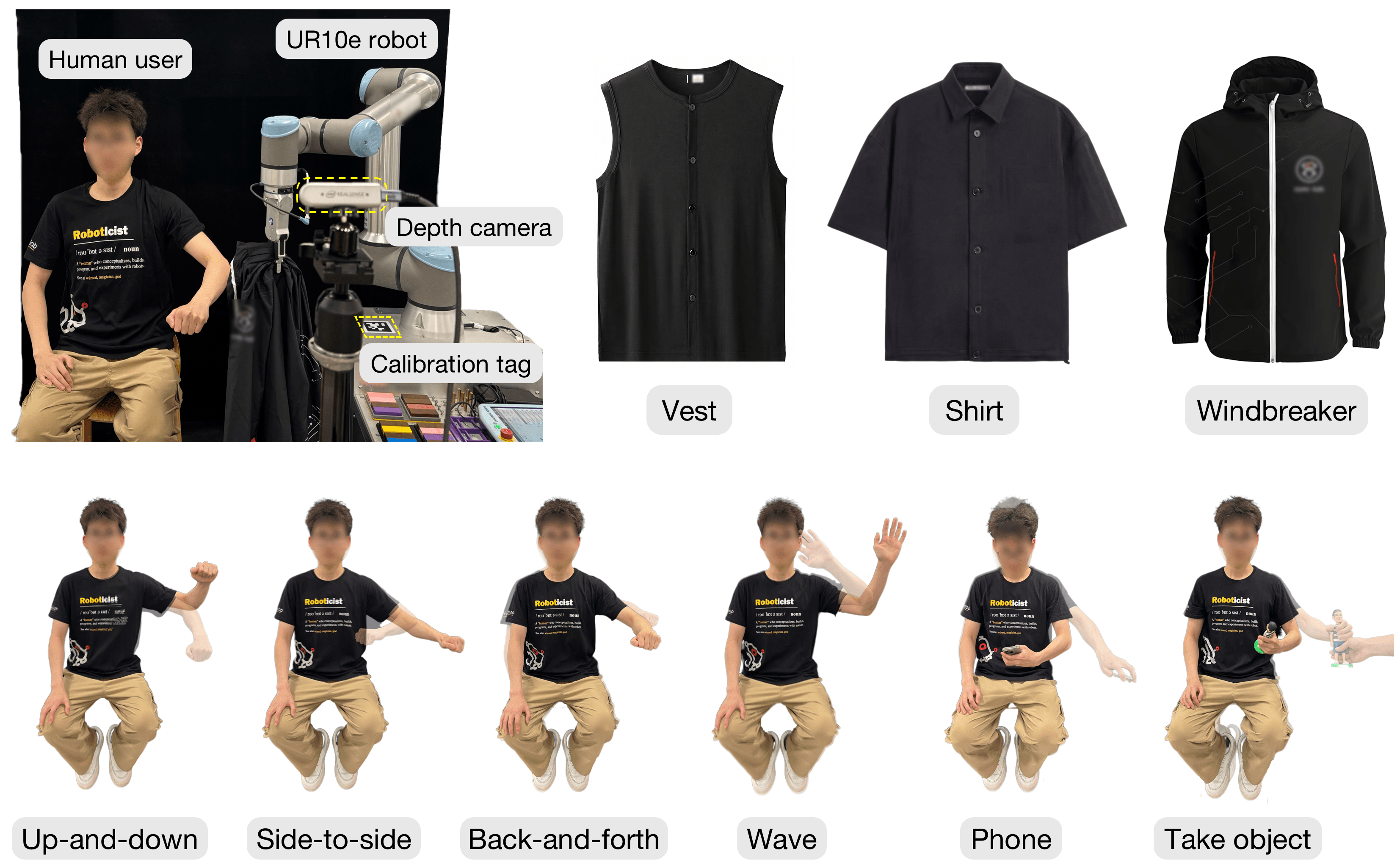}
    \caption{\textit{Top left}: Real-world experimental setup. \textit{Top right}: garments with different sleeve lengths. \textit{Bottom}: motion patterns performed by the participants.}
    \label{fig:7}
\end{figure}

Fig.~\ref{fig:7} shows the real-world experimental setup\footnote{This study was approved by the institutional ethics review board. Approval details are omitted for anonymous review and will be provided in the final version.}. We use a UR10e robot for dressing, with the participant seated to its left. A RealSense D435i camera is mounted above and in front of the participant to capture visual observations. The garments include a vest, a shirt, and a windbreaker. The human study comprises a non-scripted motion test for reactive adaptation and a canonical motion test under representative daily arm-motion patterns. Simulation and real-world experiments use independently collected datasets and separately trained policies, with no sim-to-real transfer.

\textbf{Canonical motion test.} In this test, we evaluate all benchmarks defined in Sec.~\ref{sec:numerical_simulation} under six canonical arm-motion patterns: up-and-down, side-to-side, back-and-forth, waving, phone use, and object receiving. During data collection, we use a 3Dconnexion SpaceMouse for teleoperation and collect 210 expert demonstrations. These demonstrations do not involve arm motion; all trajectories are collected from static subjects with varied arm configurations. To reduce contact risk during execution, we set safety thresholds of $20~\mathrm{kg\cdot m/s}$ for joint momentum and $120~\mathrm{N}$ for tool force.

\begin{figure*}[t]
    \centering
    \includegraphics[width=0.95\textwidth, keepaspectratio]{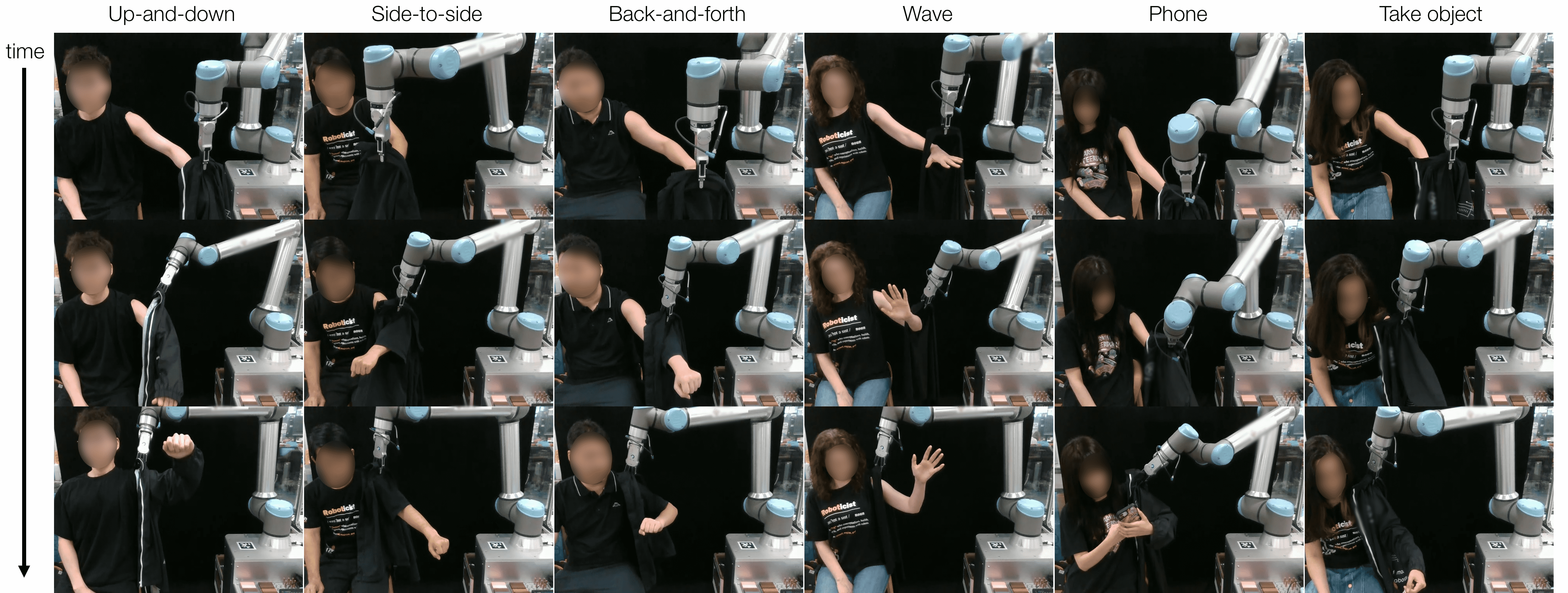}
    \caption{Snapshots of real-world dressing trials using the proposed method with different participants. Participants select varied initial arm configurations, including arm orientations and shoulder and elbow joint angles. The policy first aligns the garment opening with the hand and then adaptively updates the execution trajectory in response to subsequent arm motion.}
    \label{fig:8}
\end{figure*}

We recruit nine participants, including three females and six males, with heights ranging from 162 cm to 185 cm. Considering potential physical fatigue, each participant performs 28 trials with a random assignment of two garment types and two arm-motion patterns, resulting in 36 trials per method with balanced assignments across garments and motion patterns. For each assigned condition, the same motion sequence is repeated across all methods. Each trial consists of two stages: an initial static insertion phase and a subsequent canonical motion phase. The trial is terminated when invalid garment--body contact occurs or when the robot remains stalled for more than 5 seconds. We then measure the covered arm region to compute the dressing ratio. Example demonstrations of dressing with the proposed method are shown in Fig.~\ref{fig:8}.

\begin{figure}[!t]
    \centering
    \includegraphics[width=0.90\columnwidth, keepaspectratio]{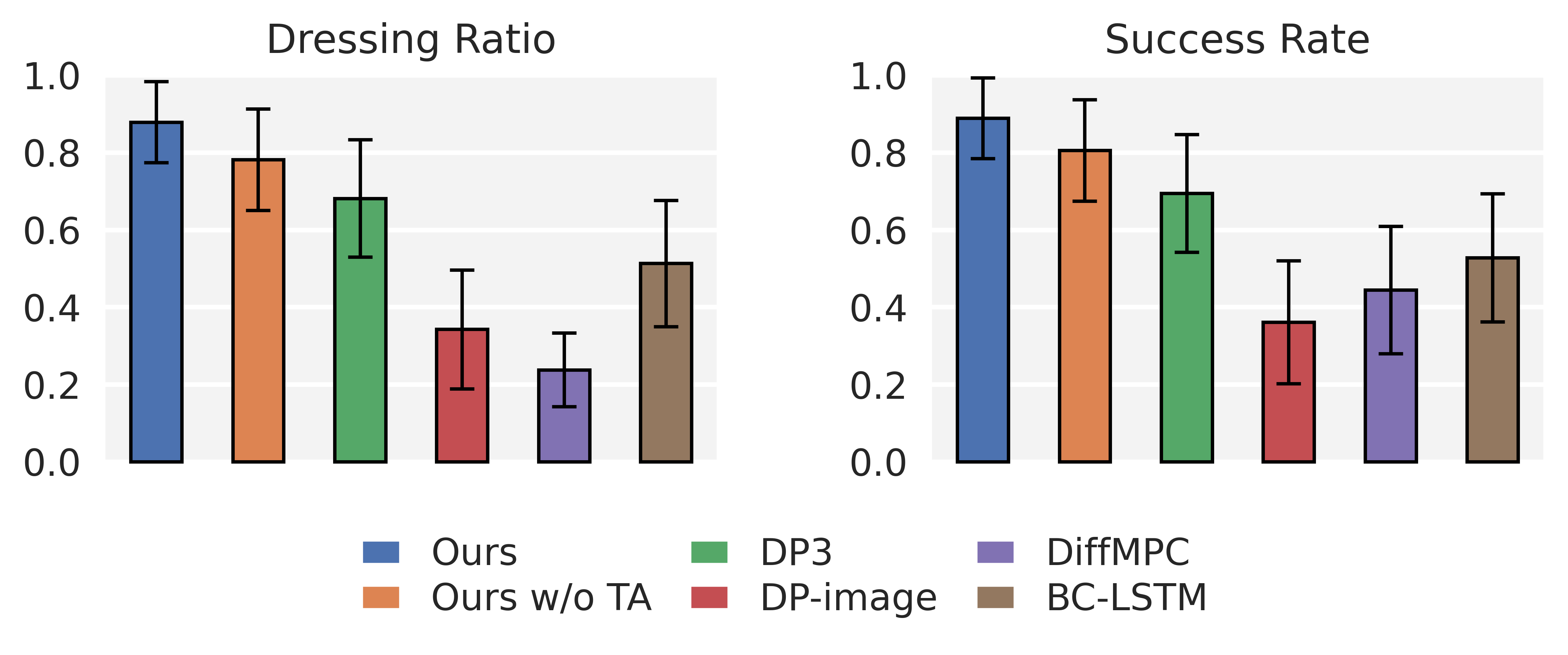}
    \caption{Dressing outcomes of the real-world human study. \textit{Left}: dressing ratio. \textit{Right}: sleeve insertion success rate.}
    \label{fig:9}
\end{figure}

\begin{table}[!t]
\centering
\caption{Dressing Ratio across Garments}
\label{tab:1}
\scriptsize
\setlength{\tabcolsep}{12pt}
\renewcommand{\arraystretch}{1.0}
\begin{tabular}{lccc}
\toprule
\textbf{Method} 
& \textbf{Vest} 
& \textbf{Shirt} 
& \textbf{Windbreaker} \\
\midrule
Ours        & 0.92 & 0.89 & \textbf{0.83} \\
Ours w/o TA & \textbf{0.93} & \textbf{0.92} & 0.50 \\
DP3         & 0.79  & 0.58 & 0.67 \\
DP-image    & 0.45  & 0.33 & 0.25 \\
Diff-MPC    & 0.14  & 0.30 & 0.27 \\
BC-LSTM     & 0.58  & 0.50 & 0.46 \\

\bottomrule
\end{tabular}
\end{table}

\begin{table}[!t]
\centering
\caption{Dressing Ratio across Motion Patterns}
\label{tab:2}
\renewcommand{\arraystretch}{1.05}
\resizebox{\columnwidth}{!}{
\begin{tabular}{l@{\hspace{0pt}}cccccc}
\toprule
\textbf{Method} 
& \textbf{Up-down} 
& \textbf{Side2side} 
& \textbf{Back-forth} 
& \textbf{Wave} 
& \textbf{Phone} 
& \textbf{Object} \\ 
\midrule
Ours        & \textbf{1.00} & \textbf{1.00} & \textbf{1.00} & \textbf{1.00} & \textbf{0.77} & 0.50 \\
Ours w/o TA & 0.83 & 0.83 & 0.83 & 0.83 & 0.76 & \textbf{0.60} \\
DP3         & 0.50 & 0.83 & 0.50 & \textbf{1.00} & 0.75 & 0.50 \\
DP-image    & 0.31 & 0.17 & 0.50 & 0.50 & 0.33 & 0.25 \\
Diff-MPC    & 0.29 & 0.42 & 0.20 & 0.20 & 0.12 & 0.20 \\
BC-LSTM     & 1.00 & 0.17 & 0.50 & 0.92 & 0.49 & 0.00 \\
\bottomrule
\end{tabular}}
\end{table}

\begin{figure}[t]
    \centering
    \includegraphics[width=0.95\columnwidth, keepaspectratio]{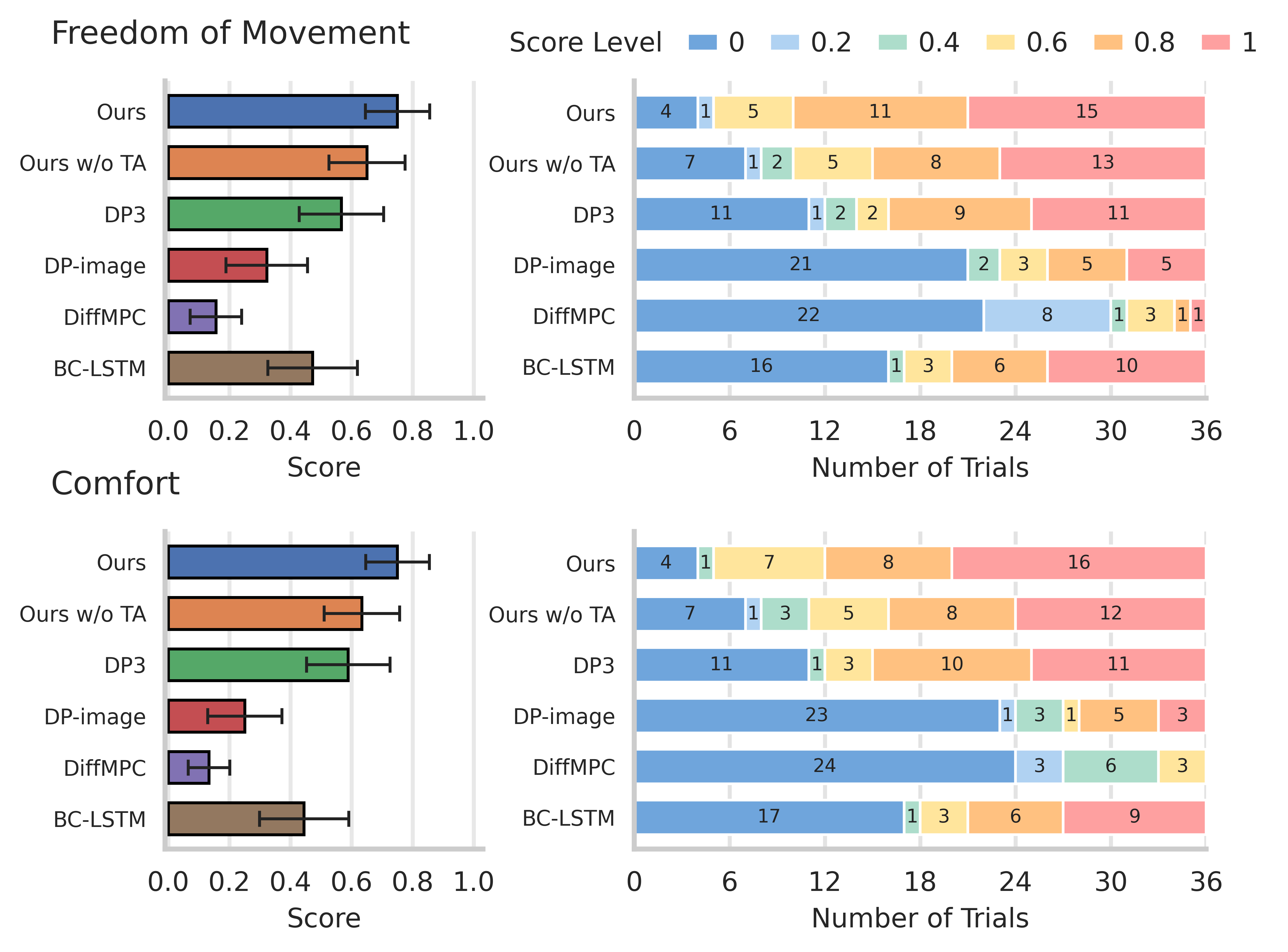}
    \caption{Subjective rating results from the real-world human study. \textit{Top}: freedom of movement. \textit{Bottom}: comfort. \textit{Left}: box plots of the rating results. \textit{Right}: full distributions of participant ratings.}
    \label{fig:10}
\end{figure}

\begin{figure}[t]
    \centering
    \includegraphics[width=0.92\columnwidth, keepaspectratio]{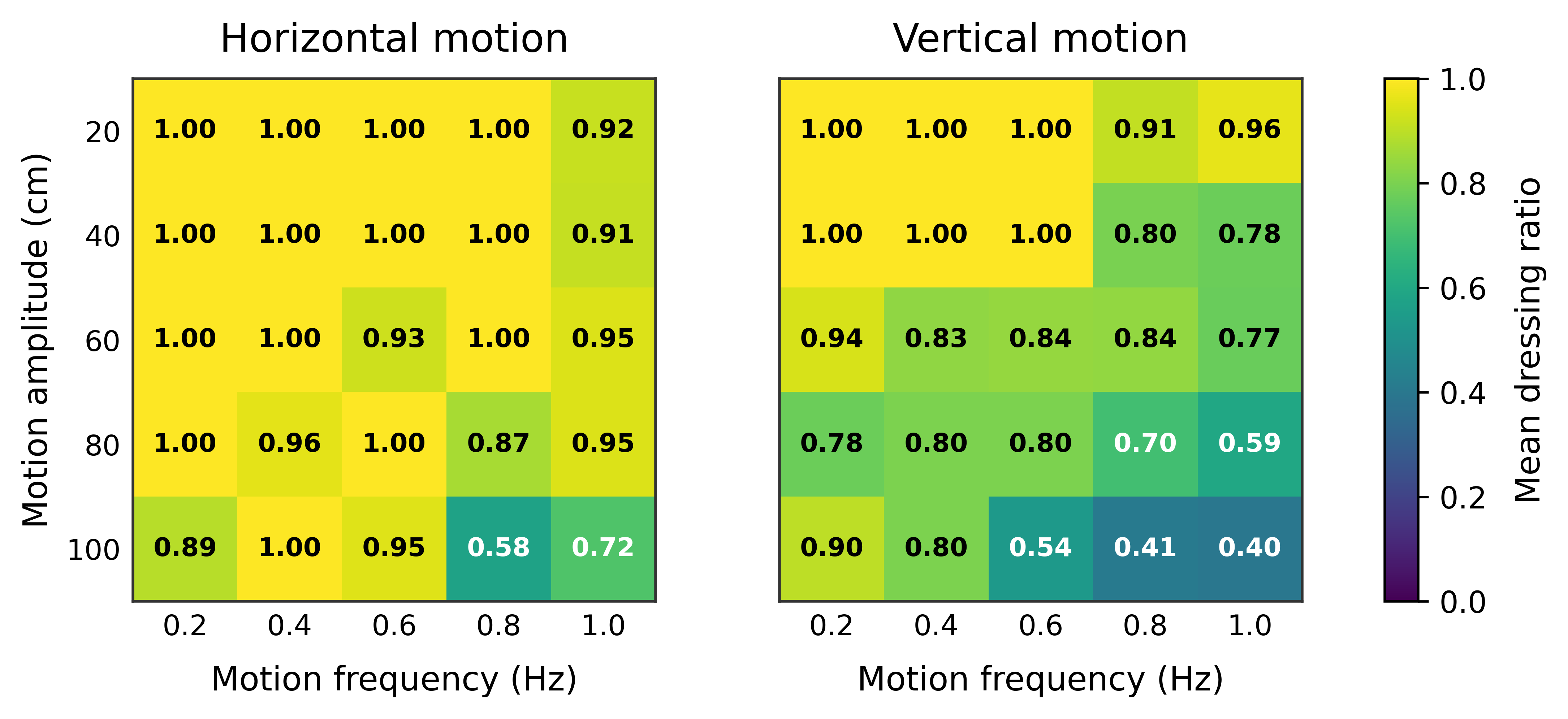}
    \caption{Results of dressing with different arm motion amplitudes and speeds. \textit{Left}: horizontal motion. \textit{Right}: vertical motion.}
    \label{fig:11}
\end{figure}

\textbf{Subjective Ratings.} Two subjective ratings are assessed: freedom of movement and comfort. Both criteria are measured on a six-level ordinal scale, with scores uniformly assigned from 1.0 to 0.0 in increments of 0.2, where higher scores indicate better performance. Freedom of movement evaluates the degree to which the subject's arm motion is constrained during dressing, as restrictions may arise from robot motion or garment pulling. A policy with better awareness of human motion is expected to achieve a higher score. Comfort evaluates the perceived contact quality during dressing, mainly reflecting the contact forces between the human body, the gripper, and the garment. More accurate dressing trajectories generally lead to smoother and more comfortable contact.

Fig.~\ref{fig:9} reports the real-world dressing outcomes. Tables~\ref{tab:1} and \ref{tab:2} summarize the dressing ratios of all methods across garment types and motion patterns, and Fig.~\ref{fig:10} reports the subjective ratings of freedom of movement and comfort. From these results, we draw the following conclusions. Consistent with the simulation results, our method outperforms all baselines in dressing progress, tolerance to human motion, and user-perceived comfort. This demonstrates that our policy successfully deployed in dynamic dressing scenarios that mostly have distribution shift, achieving a dressing ratio of $0.88\pm0.105$ and a success rate of $89\%\pm10.4\%$ over all 252 trials. Moreover, experimental results indicates that the trajectory adaptation module adaptively adjusts the demonstrated trajectories to arm motion and exhibits compliant-like behavior in most trials. As reflected by the participants, this reduces the perceived physical burden during dressing. In addition, Fig.~\ref{fig:8} provides examples in which the learned policy exhibits dressing behaviors conditioned on human morphology and pose rather than a fixed motion pattern, including backward tilting of the gripper when the shoulder is raised in the up-and-down motion and oblique gripper adjustment for upward draping over the shoulder during phone use. In contrast, the baselines reveal different limitations. Behavior cloning methods tend to learn averaged motions, while Diff-MPC is affected by the sim-to-real gap in garment dynamics and usually produces excessive elbow contact during full-sequence execution. DP3 achieves the closest performance to our method among the baselines, but still lacks effective generalization to dynamic human motion.

\textbf{Capability test.} To evaluate the motion-tolerance envelope of the proposed method, we conduct an additional real-world test by sweeping arm-motion amplitude and frequency in two canonical motion planes. Specifically, we evaluate horizontal and vertical arm swings across five amplitudes of 20--100 cm and five frequencies of 0.2--1.0 Hz, resulting in 150 trials. Fig.~\ref{fig:11} reports a clear performance degradation appears in the high-intensity region: when the arm-swing amplitude reaches 80--100 cm and the frequency reaches 0.8--1.0 Hz, the policy's ability to generate motion-compatible actions starts to degrade, leading to reduced dressing progress.

\textbf{Runtime.} In our experiments, all runtime measurements are obtained on a NVIDIA RTX 5070 Laptop GPU. Following an asynchronous inference stack similar to \cite{capuanoRobotLearningTutorial2025}, DDIM completes visual processing and action-chunk inference in less than 50 ms on average. In addition, for each frame, a 96 points region of interest $\mathcal{R}_t$ is sampled from the 1024 arm points $\mathcal{U}_t$ and registered to $\mathcal{U}_{t+1}$; the lightweight \texttt{small\_gicp} \cite{koideSmall_gicpEfficientParallel2024} implementation completes this registration in less than 10 ms on average. This enables the overall action-generation pipeline to run online with reactive updates.

\textbf{Limitations.} One limitation is that arm motion before garment--body interaction may perturb opening alignment with the hand and compromise sleeve insertion. This partly arises from the mismatch between limited control frequency and variable human motion. Second, mapping arm motion to end-effector offsets inherently introduces irregular displacement. The bounded correction in action projection must balance jitter suppression with the representation of the underlying arm-motion trend, which inevitably limits the tightness with which rapid or large-amplitude human motions can be tracked.

\vspace{-1.0em}

\section{Conclusion}
In this letter, we propose a new visuomotor policy for robot-assisted dressing that accommodates human arm motion. A diffusion policy tailored to human--garment interaction is proposed for learning dressing actions from expert demonstrations conditioned on static arm configurations. We further introduce an arm-centric diffused scalar field based on PDE surface diffusion to characterize the local axial distribution of the arm. Based on this representation, a point-cloud sampling and registration method is developed to approximate arm motion through coordinate transformation and project actions online for motion-aligned trajectory adaptation. We evaluate the proposed method in simulation and real-world experiments involving nine participants, three garment types and six arm-motion patterns. The results show that our method generalizes effectively to novel arm configurations and user motions, outperforming six baselines in dressing progress, sleeve insertion success, freedom of movement, and user comfort.

\appendices


\bibliography{bio.bib} 
\bibliographystyle{IEEEtran}

\end{document}